\pdfoutput=1
\documentclass[10pt,twocolumn,letterpaper]{article}
\usepackage[pagenumbers]{cvpr}
\usepackage{microtype}
\usepackage{array,tabularx}
\usepackage{iftex}
\ifXeTeX
  \usepackage{fontspec}
\fi
\newcommand{\method}{RecastAgent}

\definecolor{cvprblue}{rgb}{0.21,0.49,0.74}
\usepackage[pagebackref,breaklinks,colorlinks,allcolors=cvprblue,hypertexnames=false]{hyperref}
\hypersetup{pdfauthor={Shiying Zhang, Yuxia Zhu, Huaze Xue, Damon Dang, Yizhen Wu, Yijie Li},pdftitle={What to Change, What to Keep: Experience-Evolving Agents for Video Recasting},pdfsubject={},pdfkeywords={}}
\def\confName{CVPR}
\def\confYear{2027}

\title{What to Change, What to Keep: Experience-Evolving Agents for Video Recasting}
\author{%
Shiying Zhang$^{2,*}$\quad Yuxia Zhu$^{3,*}$\quad Huaze Xue$^{4,*}$\\
Damon Dang$^{2}$\quad Yizhen Wu$^{2}$\quad Yijie Li$^{1,\dagger}$\\[3pt]
{\normalsize $^{1}$MoE Key Laboratory of Interdisciplinary Research of Computation and Economics,}\\
{\normalsize Shanghai University of Finance and Economics}\\
{\normalsize $^{2}$Tongji University\quad $^{3}$Peking University\quad $^{4}$Tsinghua University}\\[3pt]
{\small $^{*}$These authors contributed equally to this work.\quad $^{\dagger}$Corresponding author: Yijie Li.}\\[3pt]
{\small
\href{https://github.com/Zzzyxii/RecastAgent}{\raisebox{-1.5pt}{\includegraphics[height=1.05em]{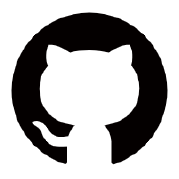}}\hspace{3pt}\textbf{Code}}\qquad
\href{https://zzzyxii.github.io/RecastAgent/}{\raisebox{-1.5pt}{\includegraphics[height=1.05em]{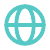}}\hspace{3pt}\textbf{Website}}}%
}

\begin{document}
\maketitle
\begin{abstract}
Video recasting aims to transform the entities and environments of existing videos into new visual worlds while preserving their essential actions and interactions. However, achieving coherent transformations remains challenging, as coordinated edits can disrupt the original event structure. In this paper, we introduce \method{}, an experience-guided agentic framework for source-conditioned video recasting that operates under frozen model weights. Our framework employs an Experience-Guided Planner to choose edit prompts and input video states under qualitative visual critique, allowing the agent to build upon partial progress or backtrack to preserve source structure. Across tasks, an Experience Accumulation module distills completed trajectories into conditional, evidence-linked principles that record both an editing strategy and its scope of validity. We additionally construct a benchmark of 67 source videos with 134 paired-specification tasks across six creative categories for evaluation. Experiments demonstrate that \method{} outperforms single-pass and agentic rewriting baselines across Bernini and Ditto. Furthermore, accumulated experience improves performance on held-out test videos and transfers zero-shot to external benchmarks like FiVE-Bench and VE-Bench.
\end{abstract}

\section{Introduction}
\label{sec:intro}

Videos are widely used in storytelling, advertising, and creative media, where creators often repurpose existing footage through editing and visual transformation to convey new narratives. However, seemingly simple creative requests—such as reimagining an iconic action within a radically different artistic medium or world—are challenging to realize because they require coordinated changes to multiple visual elements while preserving temporal consistency. We study this task as \emph{video recasting}, which aims to reconstruct and reinterpret the entities and environments of existing footage within a new video world, while preserving the underlying event structure, including essential actions, interactions, and spatiotemporal relationships. In essence, video recasting changes the world in which an event unfolds, rather than the event itself.

Prior work has explored video editing from two main directions. Diffusion-based and instruction-driven editors enable diverse visual transformations \cite{tuneavideo,fatezero,tokenflow,vace,ditto,bernini}, while agent-based approaches employ planning and visual feedback to translate user intent into editing operations \cite{videodirectorgpt,videostudio,genmac,vista,editduet,aurora}. However, these approaches still face challenges in complex video recasting, which requires coherently transforming multiple entities and their environments while preserving the original actions, interactions, and spatial relationships. These requirements are closely coupled: transformed entities must not only conform to the target world but also remain consistent with the original motion and surrounding environment. As a result, existing methods may produce incomplete transformations, incorrect entity–attribute associations, or unintended changes to the original actions and interactions. Achieving coherent world transformations while preserving the underlying structure of the source video therefore remains challenging.

\begin{figure*}[t]
  \centering
  \includegraphics[width=\textwidth]{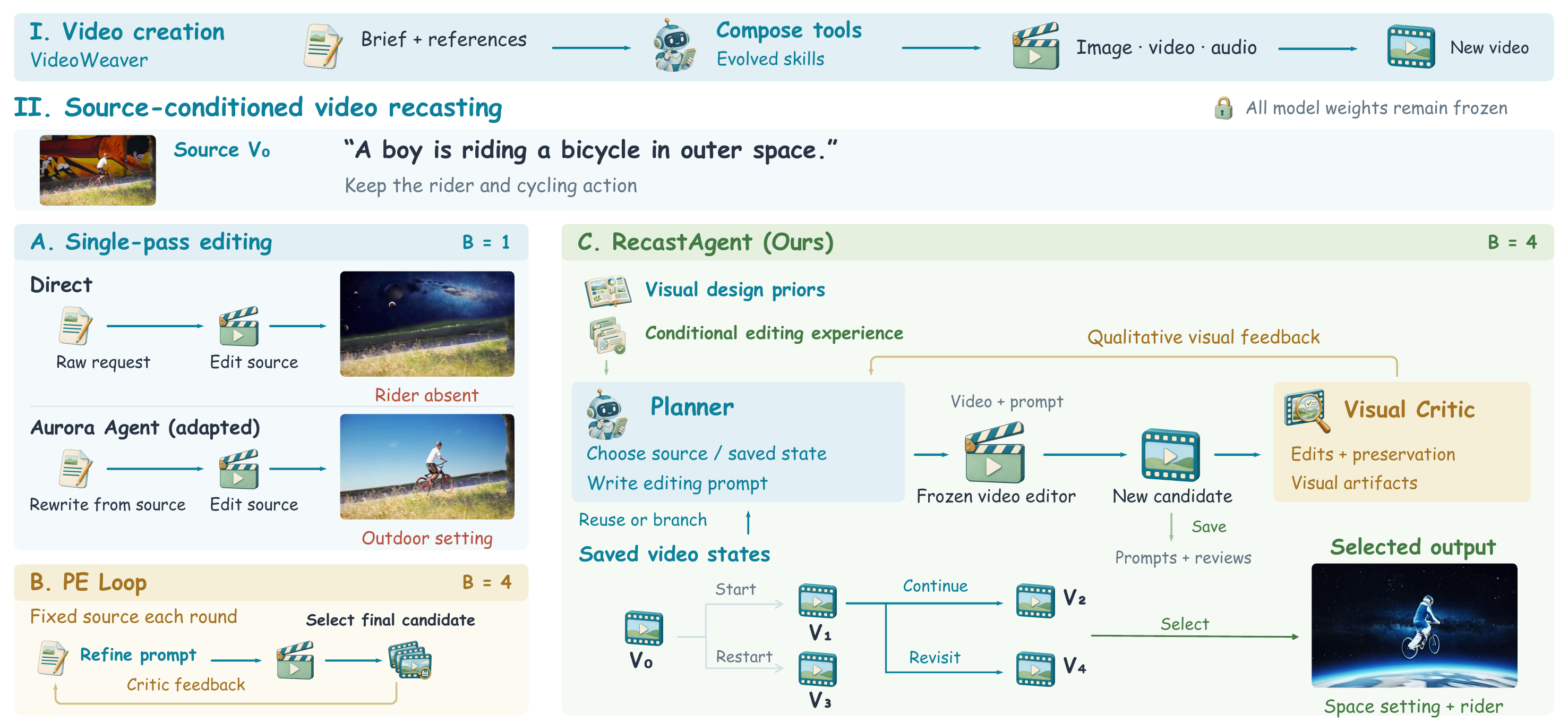}
  \caption{Agentic video creation and source-conditioned recasting.
  VideoWeaver~\cite{videoweaver} composes generation tools. Direct and adapted
  Aurora~\cite{aurora} each edit once; PE Loop revises prompts while returning
  to the source. \method{} chooses both video states and prompts under an
  independent Critic, consulting frozen design references and experience.
  The branch diagram is schematic; PE was not run on the illustrated example.
  All model weights remain frozen.}
  \label{fig:overview}
\end{figure*}

A promising direction is to construct precise multimodal prompts—combining text and visual references—to specify target transformations while preserving original actions and interactions. Iterative refinement via visual feedback can further improve results through test-time scaling, yet this process requires extensive trial and error, and gained insights typically remain isolated to single tasks. This leads to redundant exploration and motivates us to distill editing trajectories into reusable knowledge for future recasting decisions. 

To this end, we introduce \textbf{\method{}}, a training-free agentic framework that accumulates and reuses experience across iterative video recasting trajectories. It comprises three core components: (1) an \textit{Experience-Guided Planner}, which leverages visual references and prior experience to formulate prompts and select the input video state (source or intermediate) at each step; (2) a \textit{Visual Critic}, which evaluates outputs against the user request to guide refinement and final selection; and (3) an \textit{Experience Accumulation} module, which distills completed trajectories into reusable strategies—complete with applicability conditions and visual evidence—stored in an external library to guide unseen tasks.

To evaluate the effectiveness of \method{} and distinguish the contributions of within-task planning and cross-task experience accumulation, we construct a video recasting benchmark comprising 67 source videos across six creative categories, complementing existing video generation and editing benchmarks~\cite{vbench,evalcrafter,ivebench,five,vebench}. Experiments with two video editors, Bernini~\cite{bernini} and Ditto~\cite{ditto}, demonstrate that \method{} achieves superior results compared to existing baselines. Furthermore, accumulated experience improves performance on testset and enables zero-shot generalization to FiVE-Bench~\cite{five} and VE-Bench~\cite{vebench}.

Our main contributions are summarized as follows:
\begin{itemize}
    \item We introduce \emph{video recasting} as a structured video adaptation task and propose \method{}, an agentic framework that coordinates iterative editing decisions through planning and visual feedback.
    \item We develop an experience accumulation mechanism that distills completed editing trajectories into reusable, evidence-grounded knowledge, enabling cross-task improvement without model training.
    \item We establish a video recasting benchmark with paired creative instructions and a controlled evaluation protocol. Extensive experiments demonstrate the effectiveness of our framework and the transferability of accumulated experience across unseen videos and benchmarks.

\end{itemize}

\section{Related work}\label{sec:related}
\paragraph{Video editing and agentic creation.}
One-shot and training-free adaptation support video edits
\cite{tuneavideo,text2videozero}; inversion, attention manipulation, and feature
propagation improve correspondence
\cite{fatezero,videop2p,pix2video,tokenflow}. Temporal constraints
\cite{flatten,vidtome,rave} and structural or key-frame conditioning
\cite{rerender,codef,controlvideo,anyv2v} help preserve content. Unified editors
include VACE, Ditto, and Bernini~\cite{vace,ditto,bernini}. We use editors as
frozen black-box tools rather than modify their weights.
VideoDirectorGPT and VideoStudio plan multi-scene generation
\cite{videodirectorgpt,videostudio}; GenMAC and VISTA use collaboration and
feedback~\cite{genmac,vista}. Rewriting Video, Aurora, and EditDuet address
video reauthoring or tool-using editing~\cite{rewriting,aurora,editduet}.
\method{} additionally considers saved video states and cross-task advice.

\paragraph{Self-improving agents.}
ReAct, Self-Refine, and CRITIC study reasoning and refinement
\cite{react,selfrefine,critic}. Reflexion and ExpeL retain prior lessons
\cite{reflexion,expel}, while Generative Agents and MemGPT maintain persistent
memory~\cite{generativeagents,memgpt}. Voyager, SkillWeaver, and Agentic Context
Engineering accumulate executable or contextual knowledge
\cite{voyager,skillweaver,ace}. Visual-domain approaches include VideoWeaver,
CineForge, and Video-RSI~\cite{videoweaver,cineforge,videorsi}; PAST-Bench and
MetaSkill-Evolve examine experience-driven improvement
\cite{pastbench,metaskillevolve}. Our reusable unit is conditional editing
advice with trajectory evidence, distinct from current-task Critic feedback.

\paragraph{Benchmarks.}
Video-generation benchmarks measure perceptual, semantic, temporal, and
compositional quality~\cite{vbench,evalcrafter,fetv,vbench2,t2vcompbench,genaibench}.
IVEBench, FiVE-Bench, and VE-Bench evaluate instruction-guided editing
\cite{ivebench,five,vebench}. Our paired-specification benchmark separately
assesses request completion, required source correspondence, and unrequested
damage.

\section{RecastAgent}\label{sec:method}
\begin{figure*}[t]
  \centering
  \includegraphics[width=\textwidth]{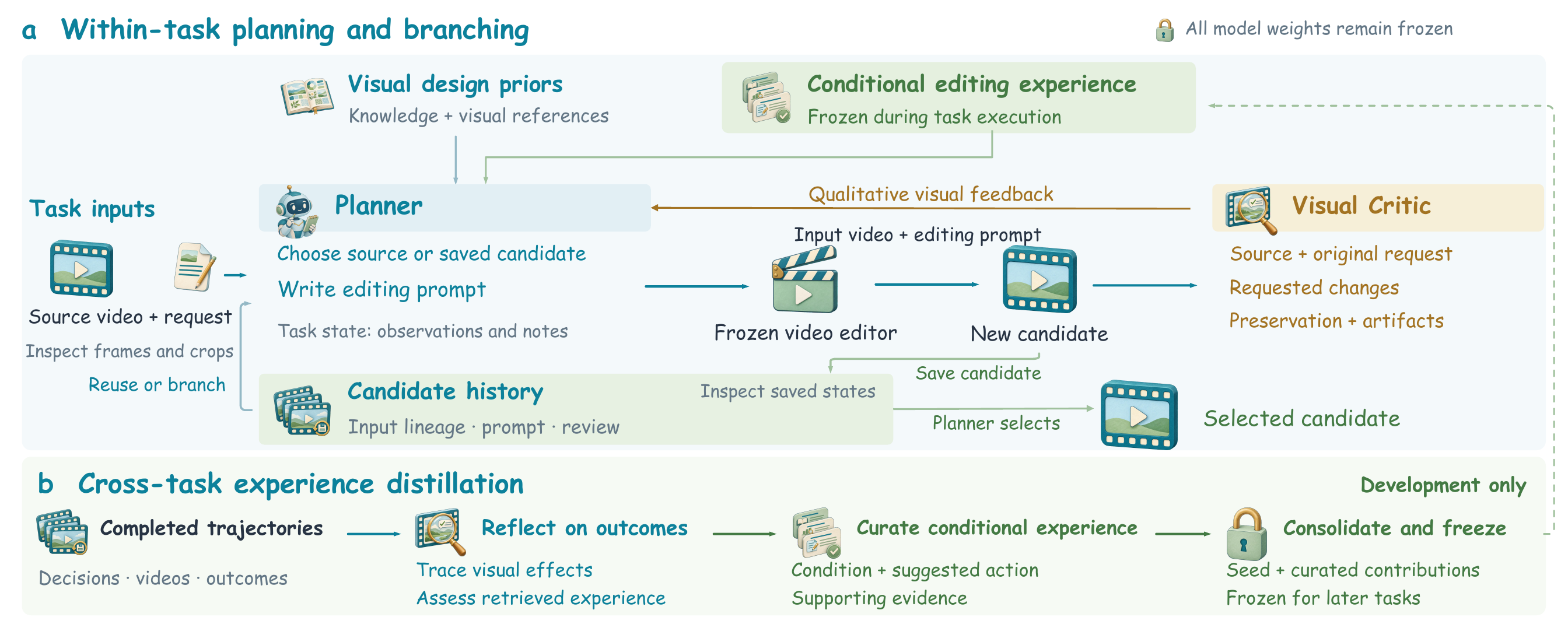}
  \caption{\method{} workflow. Source observations and frozen references feed
  the Planner. It chooses an input and prompt, records successful candidates,
  and receives independent visual critique. The total submitted-generation
  budget is four, not four per action. A delivery is selected from saved
  candidates. Only completed development trajectories supply experience
  contributions; the consolidated library is frozen before held-out tests.}
  \label{fig:method}
\end{figure*}

Given a source video $V_0$ and request $u$, the task is to construct a new
visual world while retaining the actions and relations that the request still
requires. We write
\begin{equation}
  \hat V=\mathcal A(V_0,u;\mathcal K,E,G_\theta),
  \label{eq:recasting-interface}
\end{equation}
where $\mathcal K$ contains design references, $E$ is a fixed experience
library, and $G_\theta$ is the frozen editor. Neither external knowledge source
is an observation of the current generated result. Figure~\ref{fig:method}
separates grounding, online editing, and development-only accumulation.

\subsection{Experience-guided planning}\label{sec:planning}
At attempt $t$, the Planner chooses an input $X_t$ and prompt $p_t$:
\begin{equation}
  (X_t,p_t)=\pi_\phi(V_0,u,M_t,H_{t-1};\mathcal K,E).
  \label{eq:planner}
\end{equation}
The planner weights $\phi$ are frozen. $M_t$ contains current observations and
working notes; $H_{t-1}$ records earlier inputs, prompts, outputs, failures,
and Critic feedback. These records evolve within a run; $\mathcal K$ and $E$
do not. $X_t$ must be the source or a successfully saved candidate from an
earlier attempt. Only the source is available at the first attempt.

\paragraph{Grounding and retrieval.}
The Planner distinguishes requested transformations from source roles,
actions, and relations that should remain recognizable. Recasting a cafe
writer as a pirate captain, for example, changes setting and appearance
while preserving writing unless the request authorizes another activity.
Observation tools return native frames, intervals, and spatial crops.
Retrieval tools read design references and experience entries; record tools
inspect provenance and retain notes. These actions do not consume generation
attempts. Supplementary~A illustrates the action space.
Design references describe target-world appearance and materials. An
experience entry instead records editing advice, conditions, and supporting
evidence from earlier tasks. Retrieved advice can inform a decision, but
does not prove that the current candidate succeeds.

\subsection{Feedback-driven iterative editing}\label{sec:revision}
The frozen editor receives the selected input and prompt:
\begin{equation}
  V_t=G_\theta(X_t,p_t;\xi_t),
  \label{eq:video-generation}
\end{equation}
with sampling configuration $\xi_t$ fixed by the run protocol. A successful
candidate can support a continuation; a damaged candidate may motivate a
source restart or a return to an earlier state. Submitted failures consume
budget but do not create a selectable video.

After a successful generation, the independent Visual Critic compares the
source, request, chosen input, prompt, and candidate. It can inspect native
frames and crops and returns qualitative feedback on requested changes,
preserved relations, and defects. This is current-task evidence, not prior
experience or a scalar reward. The Planner records it in the history with the
attempt provenance. An incorrect attribute binding may require a new prompt;
a lost prop or broken interaction may instead require another input state.
Once the budget is exhausted, the Planner selects a successful saved
candidate using the recorded evidence. The choice need not be the latest
generation. Interrupted and undelivered runs are reported separately
(Supplementary~C), not counted as completed four-candidate runs.

\subsection{Cross-task experience accumulation}\label{sec:experience}
For development task $k$, accumulation reviews its saved trajectory
$\mathcal T_k$, examines whether retrieved advice was applicable, and distills
conditional contributions $\Delta_k$ with evidence links. Every trajectory and
distillation pass reads the same initial library $E_0$, not updates from other
concurrent tasks. Contributions are consolidated as
\begin{equation}
 E_1=\operatorname{Consolidate}
       \left(E_0,\{\Delta_k:k\in\mathcal I_{\rm dev}\}\right).
 \label{eq:experience-update}
\end{equation}
Here $\mathcal I_{\rm dev}$ contains development tasks only. Consolidation
updates external advice, not model weights. The accumulated library $E_1$ is
frozen before testing; test-task notes and contributions are not merged into
the tested snapshot. Both Planner configurations retain the same initial
curated knowledge, so this is not a comparison against an empty library.

\section{Benchmark and evaluation}\label{sec:evaluation}
\subsection{Paired specifications and split}\label{sec:benchmark}
\begin{figure}[t]
  \centering
  \includegraphics[width=\columnwidth]{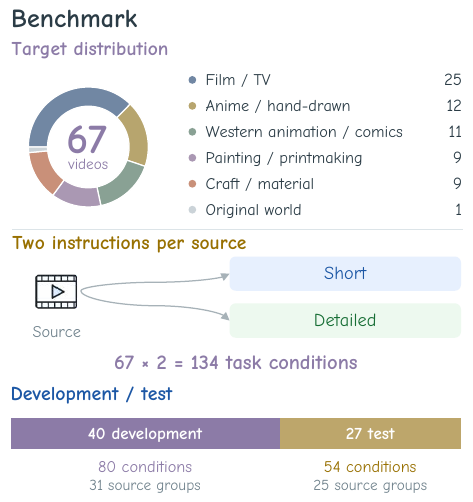}
  \caption{The 67-source benchmark. Six target categories and paired
  short/detailed instructions yield 134 conditions. Development and test
  videos are split at the related-source group level.}
  \label{fig:benchmark}
\end{figure}

\begin{figure*}[t]
  \centering
  \includegraphics[width=\textwidth]{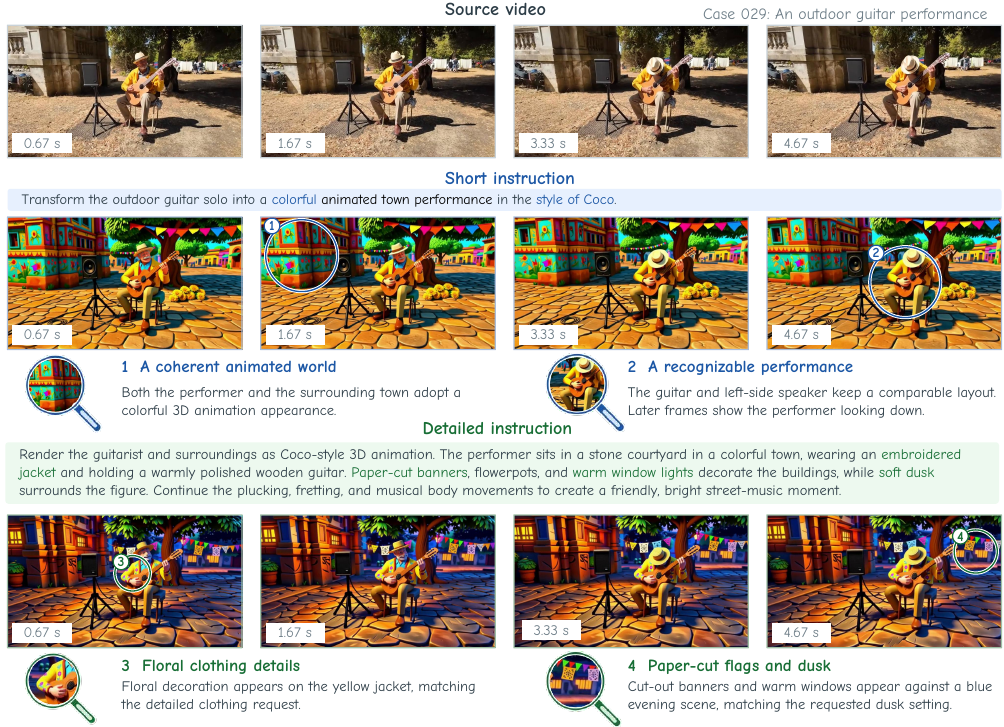}
  \caption{Paired specifications for development source 029, targeting a
  Coco-style animated performance. The detailed request adds clothing,
  courtyard, banners, warm windows, and dusk. This illustration explains
  specification granularity; it is not a selected-output comparison from the
  audited PE cohort or a verified experience-use trace. Supplementary Fig.~S2
  shows different matched examples from archived runs.}
  \label{fig:paired-specifications}
\end{figure*}

\begin{figure*}[t]
  \centering
  \includegraphics[width=\textwidth]{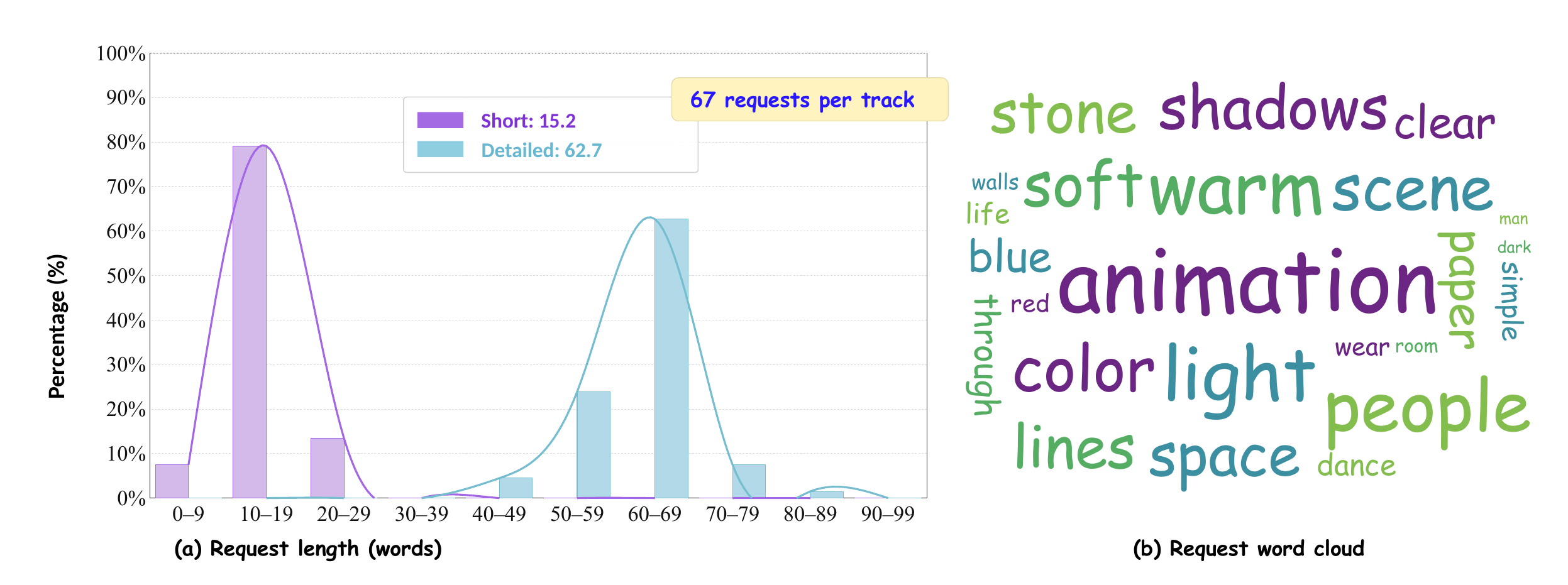}
  \caption{Descriptive statistics of the 134 frozen benchmark requests.
  Each length distribution has 67 requests in 10-word bins; curves
  only guide the eye through those frequencies. The word cloud summarizes
  content-word frequencies across both tracks after stopword removal. It
  describes request vocabulary, not output quality.}
  \label{fig:request-statistics}
\end{figure*}

Our benchmark contains 67 source videos across six creative categories
(Fig.~\ref{fig:benchmark}). Each has a short request leaving visual design
choices open and a detailed request specifying environment, attire, materials,
lighting, or interactions, giving 134 conditions. The formal frozen requests
average 15.2 and 62.7 words, respectively (Fig.~\ref{fig:request-statistics}); punctuation is excluded and internal
hyphens/apostrophes remain within words. Each output is judged against its own
request, not a substituted prompt version.
Related clips form 56 source-disjoint groups. Forty development sources
(31 groups, 80 conditions) supply experience contributions; 27 held-out sources
(25 groups, 54 conditions) do not. Figure~\ref{fig:paired-specifications}
illustrates the specification tracks for source 029. Supplementary~A presents
different matched examples from archived runs, not a causal prompt-length test.

Figure~\ref{fig:paired-specifications} retains the performer, guitar, and
musical gestures in both specifications; the detailed version adds clothing,
courtyard, decorations, and lighting. A recognizable animated performer alone
does not establish setting completion. We therefore judge completion and
correspondence separately from visual finish and across multiple moments.

\subsection{Metrics and judges}\label{sec:metrics}
Instruction diagnostics separately measure explicit-request completion,
required source correspondence, and unrequested damage. Correspondence is
request-conditioned, not literal pixel preservation. Detailed requests add
entity--attribute binding and joint satisfaction; short requests add style
realization, scene-specific design, and effective creativity.
Five-axis quality rates \emph{Target Realization} (TR), \emph{Content
Correspondence} (CP), \emph{Temporal Consistency} (TC), \emph{Visual Integrity}
(VI), and \emph{Aesthetic Coordination} (AC) from 0 to 4. Stable but unconverted
regions can receive TC credit without receiving TR credit. Full anchors and
valid-denominator rules are in Supplementary~D.

GLM-5.3-Flash and Claude-Sonnet-5.5 independently judge anonymized outputs
using source-derived frame plans and continuous-segment evidence
(Supplementary~B). Execution prompts, method identities, and online Critic
reviews are hidden. Pairwise task completion is judged twice with presentation
order swapped. We report wins, ties, and losses only for the $n$
order-consistent pairs. Remaining comparisons are unresolved, not ties.
Group-weighted score bounds retain unresolved and missing outcomes.
Five-axis means first average valid short/detailed ratings within a source,
then weight valid sources equally. Unknown ratings are excluded, never zeroed.

\section{Experiments}\label{sec:experiments}
Our experiments are designed to answer three primary research questions:

\begin{itemize}

\item \textbf{RQ1 (Agentic workflow vs.\ single-pass and prompt-only iteration):} Under frozen editors and matched generation budgets, does trajectory branching with design priors and visual critique improve video recasting over single-pass rewriting and source-only prompt iteration?
\item \textbf{RQ2 (Constraint satisfaction vs.\ open design freedom):} How effectively does \method{} execute fine-grained prescribed edits on detailed instructions and synthesize coherent visual designs on underspecified short instructions while preserving required source correspondences?
\item \textbf{RQ3 (Cross-task self-improvement and zero-shot transfer):} Does experience distilled from the agent's own development trajectories causally improve editing decisions on unseen test videos and generalize zero-shot to external public benchmarks?
\end{itemize}

\subsection{Experimental setup}\label{sec:setup}
\noindent\textbf{Frozen editors and compared baselines.} We evaluate all methods across all 134 benchmark conditions using two distinct open-source video-to-video editors, Bernini \citep{bernini} and Ditto \citep{ditto}, under identical frozen checkpoints and sampling schedules (Table~S1, Appendix~A). We compare five configurations: (1) \emph{Direct} ($B=1$), which feeds the raw user request and source video directly to the editor; (2) \emph{Aurora Agent} \citep{aurora} ($B=1$), an open-source multimodal editing agent that uses its official Qwen3-VL-8B planner and released LoRA adapter to rewrite the source video and user brief into a structured editing prompt; (3) \emph{PE Loop} ($B=4$), a multi-round prompt-engineering baseline that iteratively refines the prompt using source frames and independent Critic feedback across four generations while always re-editing \(V_0\), followed by a VLM selector; and (4) \emph{Planner (\(E_0\))} and (5) \emph{Planner (\(E_1\))} ($B=4$), which run \method{} with design priors \(\mathcal K\), dynamic state branching, and either the initial experience \(E_0\) or the self-accumulated experience snapshot \(E_1\).

\noindent\textbf{Backbone models and budget control.} To ensure a fair comparison across multi-round methods, online planning, prompt enhancement (in PE Loop), visual critique, candidate selection, and post-task reflection all use GPT-6-Luna with high reasoning effort, whereas Aurora Agent uses its official fine-tuned Qwen3-VL-8B-Instruct checkpoint with LoRA. Multi-round methods (\(B=4\)) use fixed random seeds 42--45 across generations \(t=1,\dots,4\), and single-pass methods (\(B=1\)) use seed 42. The generation budget \(B\) serves as the sole online stopping criterion. Post-task reflection on each development trajectory is capped at 4 hours, 64 model calls, and 6M tokens (at most two attempts).

\noindent\textbf{Evaluation protocol.} To address RQ1, we compare Planner (\(E_0\)) against Aurora Agent and PE Loop across all 134 conditions, and Planner (\(E_1\)) against both baselines on the 54 held-out test conditions. To address RQ2, we analyze instruction-track diagnostics across short and detailed briefs. To address RQ3, we compare Planner (\(E_1\)) directly against Planner (\(E_0\)) on the 54 held-out test conditions under identical seeds and model weights, trace how retrieved experience entries shape concrete branching and prompting decisions, and evaluate zero-shot transfer with Bernini on two standard public benchmarks, FiVE-Bench \citep{five} and VE-Bench \citep{vebench} (Table~\ref{tab:external-benchmarks}). Blind pairwise task completion serves as our primary metric, complemented by five-axis holistic quality ratings.

\begin{table*}[t]
\centering\small
\caption{Video recasting results for RecastAgent and baselines on Bernini and Ditto, evaluated on 67 source videos and 134 editing conditions.
(a) Blind pairwise task completion, with win/tie/loss rates reported for the first method in each comparison.
(b) Five-axis quality scores (0--4), macro-averaged by source using valid ratings across all conditions. Within each editor, bold and underlined values mark the best and second-best scores in each column.}
\label{tab:results}
\textbf{(a) Task completion}\par\smallskip
\begin{tabular*}{\textwidth}{@{\extracolsep{\fill}}lllcccc@{}}
\toprule
Judge & Editor & Comparison & Win (\%) & Tie (\%) & Loss (\%) & $n$ \\
\midrule
GLM & Bernini & Initial/Aurora (all) & 84.6 & 0.0 & 15.4 & 104 \\
  &  & Accumulated/Aurora (test) & 83.7 & 0.0 & 16.3 & 43 \\
  &  & Initial Final/First (all) & 56.7 & 22.2 & 21.1 & 90 \\
  &  & Accumulated Final/First (all) & 54.4 & 20.0 & 25.6 & 90 \\
  & Ditto & Initial/PE (all) & 37.3 & 10.2 & 52.5 & 59 \\
  &  & Initial/Aurora (all) & 89.6 & 1.3 & 9.1 & 77 \\
  &  & Accumulated/PE (test) & 34.6 & 11.5 & 53.8 & 26 \\
  &  & Accumulated/Aurora (test) & 85.7 & 0.0 & 14.3 & 21 \\
  &  & Initial Final/First (all) & 47.2 & 33.3 & 19.4 & 72 \\
  &  & Accumulated Final/First (all) & 61.4 & 27.1 & 11.4 & 70 \\
  &  & PE Loop/Direct (all) & 81.2 & 5.8 & 13.0 & 69 \\
Claude & Bernini & Initial/Aurora (all) & 79.8 & 0.0 & 20.2 & 114 \\
  &  & Accumulated/Aurora (test) & 84.8 & 0.0 & 15.2 & 46 \\
  &  & Initial Final/First (all) & 63.5 & 17.3 & 19.2 & 104 \\
  &  & Accumulated Final/First (all) & 53.8 & 20.2 & 26.0 & 104 \\
  & Ditto & Initial/PE (all) & 48.2 & 0.0 & 51.8 & 85 \\
  &  & Initial/Aurora (all) & 83.5 & 0.0 & 16.5 & 109 \\
  &  & Accumulated/PE (test) & 42.3 & 0.0 & 57.7 & 26 \\
  &  & Accumulated/Aurora (test) & 85.4 & 0.0 & 14.6 & 41 \\
  &  & Initial Final/First (all) & 57.1 & 19.8 & 23.1 & 91 \\
  &  & Accumulated Final/First (all) & 57.0 & 22.6 & 20.4 & 93 \\
  &  & PE Loop/Direct (all) & 74.5 & 0.0 & 25.5 & 94 \\
\bottomrule
\end{tabular*}
\par\medskip
\textbf{(b) Five-axis scores}\par\smallskip
\begin{tabular*}{\textwidth}{@{\extracolsep{\fill}}llcccccccccc@{}}
\toprule
 & & \multicolumn{5}{c}{GLM} & \multicolumn{5}{c}{Claude} \\
Editor & Method & TR & CP & TC & VI & AC & TR & CP & TC & VI & AC \\
\midrule
Bernini & Direct & 2.520 & 3.020 & \underline{3.130} & \underline{2.860} & 2.770 & 2.178 & 2.748 & 2.797 & 2.365 & 2.309 \\
 & Aurora Agent & 2.671 & 3.032 & 3.052 & 2.661 & 2.895 & 2.425 & 2.510 & 2.705 & 2.231 & 2.510 \\
 & PE Loop$^{\dagger}$ & 2.975 & \textbf{3.260} & 3.117 & \textbf{2.900} & 3.044 & 2.602 & \textbf{3.002} & 2.796 & \textbf{2.454} & 2.685 \\
 & Planner initial & \textbf{3.165} & \underline{3.226} & \textbf{3.205} & 2.798 & \underline{3.152} & \underline{2.827} & \underline{2.787} & \textbf{2.844} & \underline{2.406} & \underline{2.808} \\
 & Planner accumulated & \underline{3.137} & 3.170 & 3.099 & 2.798 & \textbf{3.202} & \textbf{2.884} & 2.739 & \underline{2.828} & 2.376 & \textbf{2.827} \\
\midrule
Ditto & Direct & 3.080 & 3.116 & 3.340 & \underline{3.049} & 3.376 & 2.924 & 2.778 & 2.978 & 2.496 & 3.022 \\
 & Aurora Agent & 2.980 & 3.037 & 3.303 & 2.867 & 3.340 & 2.796 & 2.739 & 2.947 & 2.478 & 2.969 \\
 & PE Loop & 3.314 & 3.232 & \underline{3.367} & 2.950 & 3.427 & 3.103 & \underline{3.015} & \textbf{3.015} & \textbf{2.600} & \textbf{3.145} \\
 & Planner initial & \underline{3.354} & \textbf{3.288} & 3.329 & 3.016 & \textbf{3.462} & \textbf{3.164} & \textbf{3.016} & 2.967 & 2.547 & \underline{3.133} \\
 & Planner accumulated & \textbf{3.360} & \underline{3.279} & \textbf{3.458} & \textbf{3.106} & \underline{3.445} & \underline{3.115} & 3.012 & \underline{2.995} & \underline{2.565} & 3.131 \\
\bottomrule
\end{tabular*}

\par\smallskip
{\footnotesize\raggedright
$n$ counts order-consistent pairs; \textit{test} denotes held-out
conditions. Final/First compares the output selected from four
attempts with the first generation.
$^{\dagger}$ Bernini PE Loop uses a mixed evaluation cohort.
\par}

\end{table*}

\begin{table*}[t]
\centering\small
\caption{External evaluation with Bernini on stratified
    36-case subsets of FiVE-Bench~\cite{five} and VE-Bench~\cite{vebench}.
    The accumulated library is reused without benchmark-specific updates.
    PSNR measures the unedited region in dB; FiVE-Acc and Win are percentages.
    Win compares each method against Direct. Bold/underline mark best/second
    per column. These compare complete configurations, not isolated experience effects.}
\label{tab:external-benchmarks}

\begin{tabular*}{\textwidth}{@{\extracolsep{\fill}}lccccc@{}}
\toprule
 & \multicolumn{3}{c}{FiVE-Bench~\citep{five}} & \multicolumn{2}{c}{VE-Bench~\citep{vebench}} \\
\cmidrule(lr){2-4}\cmidrule(lr){5-6}
Method & PSNR $\uparrow$ & CLIP-T $\uparrow$ & FiVE-Acc $\uparrow$ & QA Score $\uparrow$ & Win (\%) $\uparrow$ \\
\midrule
Direct & 24.61 & 26.95 & \underline{22.1} & \underline{0.067} & -- \\
Aurora Agent \citep{aurora} & \textbf{25.80} & \underline{27.27} & 18.6 & 0.037 & \underline{42.9} \\
Planner accumulated & \underline{25.51} & \textbf{27.68} & \textbf{26.4} & \textbf{0.199} & \textbf{57.1} \\
\bottomrule
\end{tabular*}
\end{table*}

\subsection{Workflow quality and refinement (RQ1)}\label{sec:overall-test}
\begin{figure*}[t]
  \centering
  \includegraphics[width=\textwidth]{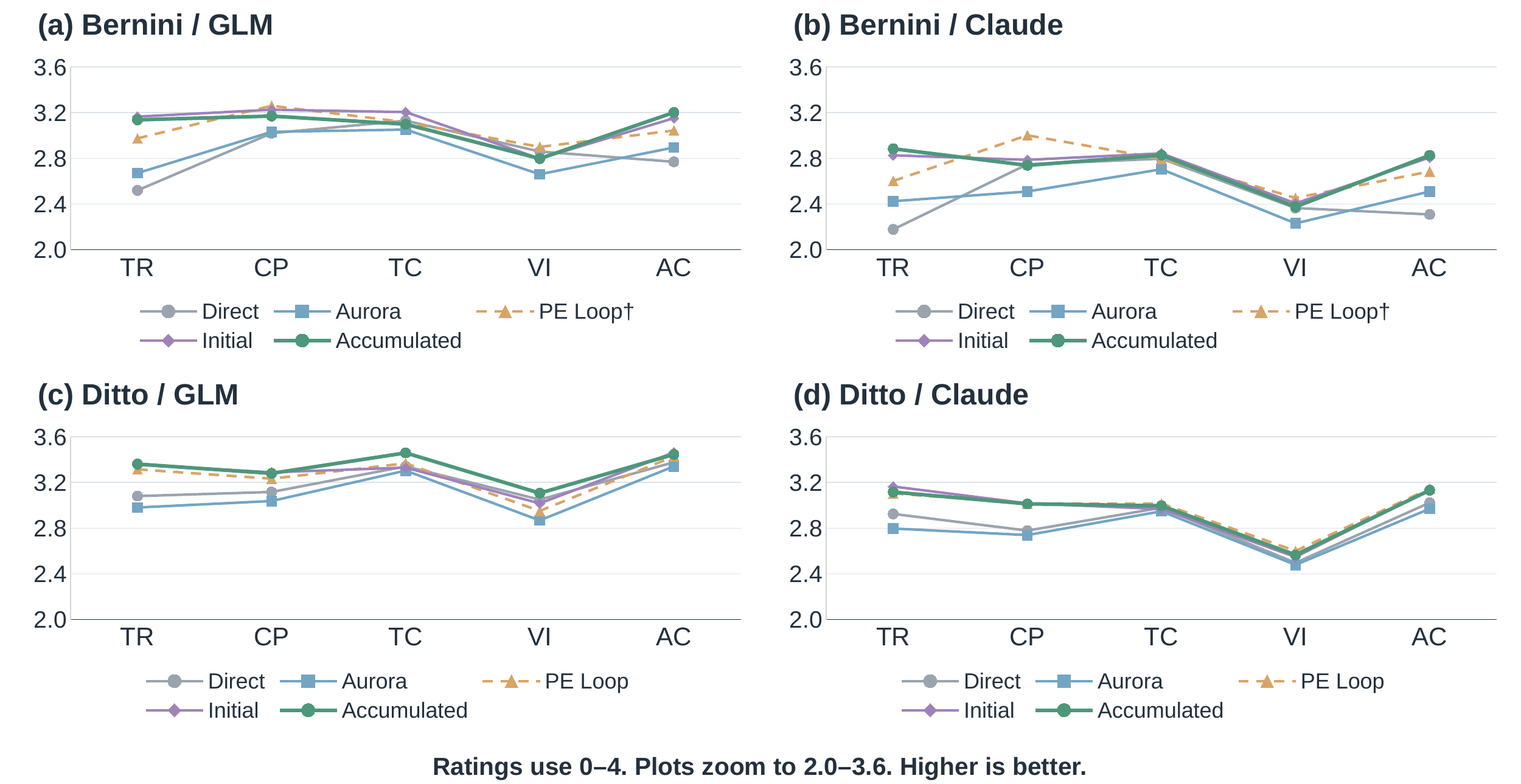}
  \caption{Measured five-axis quality, reproducing all 100 values in
  Table~\ref{tab:results}(b). Ratings use 0--4; every panel explicitly zooms
  to 2.0--3.6. Initial and accumulated denote the two Planner libraries.
  Bernini PE$^{\dagger}$ retains historical model/input strata. Full-cohort
  descriptive means do not establish held-out or causal experience gains.}
  \label{fig:results-plot}
\end{figure*}

Across both editors and judges, our complete agentic workflow substantially outperforms single-pass instruction rewriting (Fig.~\ref{fig:results-plot}, Table~\ref{tab:results}). Planner ($E_0$) achieves win rates around 80\%--90\% against Aurora across both editors, and this strong performance is maintained by Planner ($E_1$) on held-out test splits. 

\paragraph{Within-task inference-time refinement.}
Iterative planning and multi-attempt selection consistently surpass first-generation outputs. For both $E_0$ and $E_1$, the final selected delivery wins over initial attempts in the majority of resolved cases (Table~\ref{tab:results}(a)). Crucially, the Planner actively leverages its history rather than defaulting to the final iteration: an earlier candidate is chosen in over 50\% of completed runs across both editors (Supplementary~C), validating the necessity of flexible candidate retention.

\subsection{Instruction diagnostics (RQ2)}\label{sec:design-test}
Auditing the Bernini Direct baseline under complex specifications reveals a sharp contrast: while single-element completion on detailed briefs reaches moderate success, joint satisfaction across all specified constraints plummets near zero. This exposes the fundamental limitation of single-pass editing in handling multi-attribute bindings. (Detailed diagnostic breakdowns and baseline bottlenecks are cataloged in Supplementary~D).

\subsection{Held-Out Experience and Transfer (RQ3)}
Evaluating Planner ($E_1$) against Planner ($E_0$) on held-out test conditions shows judge-dependent improvements. GLM favors the accumulated library (60\%--67\% win rates), whereas Claude evaluates the two configurations near parity. While quantitative identification bounds remain inclusive, qualitative trajectory tracing confirms that retrieved experience successfully guides structural decisions—such as mapping central elements, unifying ensemble styling, and selecting superior early candidates (Supplementary~E).

\paragraph{Zero-shot external transfer.}
Evaluating the complete Planner ($E_1$) pipeline on external benchmarks (FiVE-Bench~\citep{five} and VE-Bench~\citep{vebench}) without benchmark-specific tuning demonstrates strong zero-shot generalization. Planner ($E_1$) outperforms single-pass and baseline alternatives across text-alignment, editing accuracy, and visual quality metrics (Table~\ref{tab:external-benchmarks}), confirming the broad transferability of the agentic framework.

\section{Discussion and Limitations}
\noindent\textbf{Workflow vs.\ experience gains.} Comparing Planner (\(E_0\)) against baselines evaluates the complete agentic workflow, whereas contrasting \(E_1\) with \(E_0\) isolates the causal gain of self-distilled experience under fixed weights and seeds.

\noindent\textbf{Single-batch vs.\ continual evolution.} Our frozen snapshot (\(E_0\to E_1\)) prevents test-time leakage and order dependence. Extending to continual evolution requires memory consolidation to prune redundant heuristics and resolve conflicting rules.

\noindent\textbf{Benchmark scale and calibration.} Each condition uses a single trajectory. While order-swapped VLM judging mitigates position bias, human perceptual calibration and source/exemplar attribution remain important directions for future work.

\section{Conclusion}
We presented \method{}, an experience-evolving agent for source-conditioned video recasting that unifies within-task trajectory branching with cross-task knowledge distillation under frozen weights. Specifically, \method{} jointly selects candidate video states and prompts under qualitative visual critique to balance building on partial progress against backtracking for structural fidelity, while reflecting on completed trajectories to distill conditional, evidence-linked principles. Evaluated on our 67-source (134-condition) paired-specification benchmark across Bernini and Ditto, plus zero-shot on FiVE-Bench and VE-Bench, \method{} substantially outperforms single-pass and agentic rewriting baselines, demonstrating that qualitative editing experience causally improves decisions on unseen videos.

\section*{LLM Usage Statement}
ChatGPT (OpenAI) was used for grammar correction, phrasing refinement,
and checks of manuscript consistency and references.
As components of the experimental workflow, GPT-6-Luna was used for
planning, prompt refinement, visual critique, candidate selection,
and post-task reflection. The Aurora baseline used
Qwen3-VL-8B-Instruct with its released LoRA adapter for prompt
rewriting. Bernini and the video editing model released with Ditto
were used to generate the experimental video outputs.
GLM-5.3-Flash and Claude-Sonnet-5.5 served as automated evaluators.
These uses and their configurations are described in the method
and experimental sections.
The authors designed and supervised the study, reviewed and
verified the final manuscript and reported analyses, and retain
full responsibility for the accuracy, attribution, interpretation,
and reproducibility of the work.
\par

{\small
  \bibliographystyle{ieeenat_fullname}
  \bibliography{main}
}

\clearpage

\maketitlesupplementary
\makeatletter
\setlength{\@fptop}{0pt}
\setlength{\@dblfptop}{0pt}
\makeatother
\appendix
\setcounter{figure}{0}
\renewcommand{\thefigure}{S\arabic{figure}}
\renewcommand{\theHfigure}{S\arabic{figure}}
\setcounter{table}{0}
\renewcommand{\thetable}{S\arabic{table}}
\renewcommand{\theHtable}{S\arabic{table}}
\clubpenalty=10000
\widowpenalty=10000
\raggedbottom
\section{Editing workflows and matched examples}\label{app:records}
A recasting instruction specifies what to change; its source video establishes
the action and relationships that should survive the edit. Each of the 67
sources has a short and a detailed request. The resulting 134 conditions
cover 40 development sources and 27 held-out sources. Related clips stay
in the same split: 31 related-source groups for development and 25 for test.

\begin{figure}[ht]
\centering
\includegraphics[width=\linewidth]{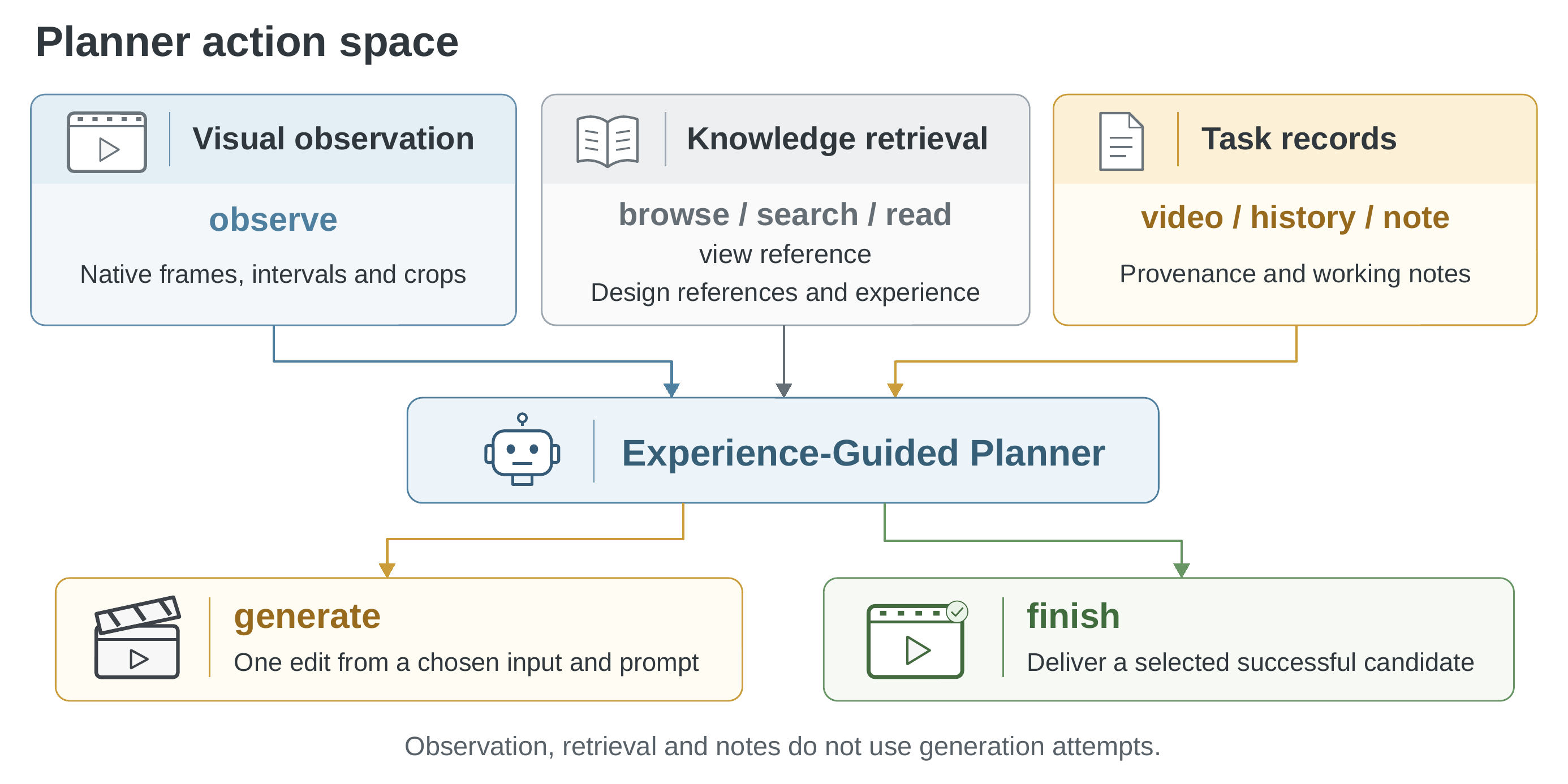}
\caption{The Planner can inspect a candidate, retrieve advice, and either
generate another edit or select a saved candidate.}
\label{fig:actions}
\end{figure}

\noindent\textbf{Where each edit starts.}
Direct uses the original request and Aurora Agent rewrites it; each generates
one video. PE Loop revises the prompt after Critic feedback, but starts each
of four edits from the original source. The Planner can also continue an
intermediate edit. Its action space (Figure~\ref{fig:actions}) therefore
separates two choices: which video to edit next and which saved version to
deliver. A submitted generation uses one attempt, including when it fails.
Observation and retrieval leave the generation budget unchanged.

\noindent\textbf{Generation settings.}
The multi-attempt methods use seeds 42--45; Direct and Aurora use seed 42.
Table~\ref{tab:renderer-settings} lists the two frozen editors' settings.
Prompt revision in PE Loop and the Planner takes place before the editor
is called. For the Bernini PE row marked $\dagger$ in main
Table~\ref{tab:results}, 117 conditions used Luna/high and 17 used Sol/high.
Of the 134 deliveries, 126 used full-span 117-frame inputs at 24 fps.
One used a source prefix with the same timing, and seven used native prefixes
of 81, 97, or 101 frames at their original rates.
\begin{table}[ht]
\centering\small
\caption{Frozen renderer settings. Both editors disable built-in prompt
enhancement and use an empty negative prompt.}
\label{tab:renderer-settings}
\begin{tabular*}{\linewidth}{@{\extracolsep{\fill}}lcc@{}}
\toprule
Setting & Bernini & Ditto \\
\midrule
Inference steps & 40 & 50 \\
Guidance & APG (v2v) & CFG 5.0 \\
Additional LoRA & None & Coefficient 1.0 \\
Network precision & bfloat16 & bfloat16 \\
VAE / tiling & float32 / default & tiled inference \\
Text capacity & 512 tokens$^{*}$ & 512 tokens \\
Maximum side & 848 pixels & 848 pixels \\
\bottomrule
\end{tabular*}
\par\smallskip\raggedright\footnotesize
$^{*}$Includes the official prefix. Ditto preserves source aspect ratio
and rounds dimensions to multiples of 16.
\end{table}

\begin{figure*}[!t]
\centering
\includegraphics[width=\textwidth,trim=0 493bp 0 0,clip]{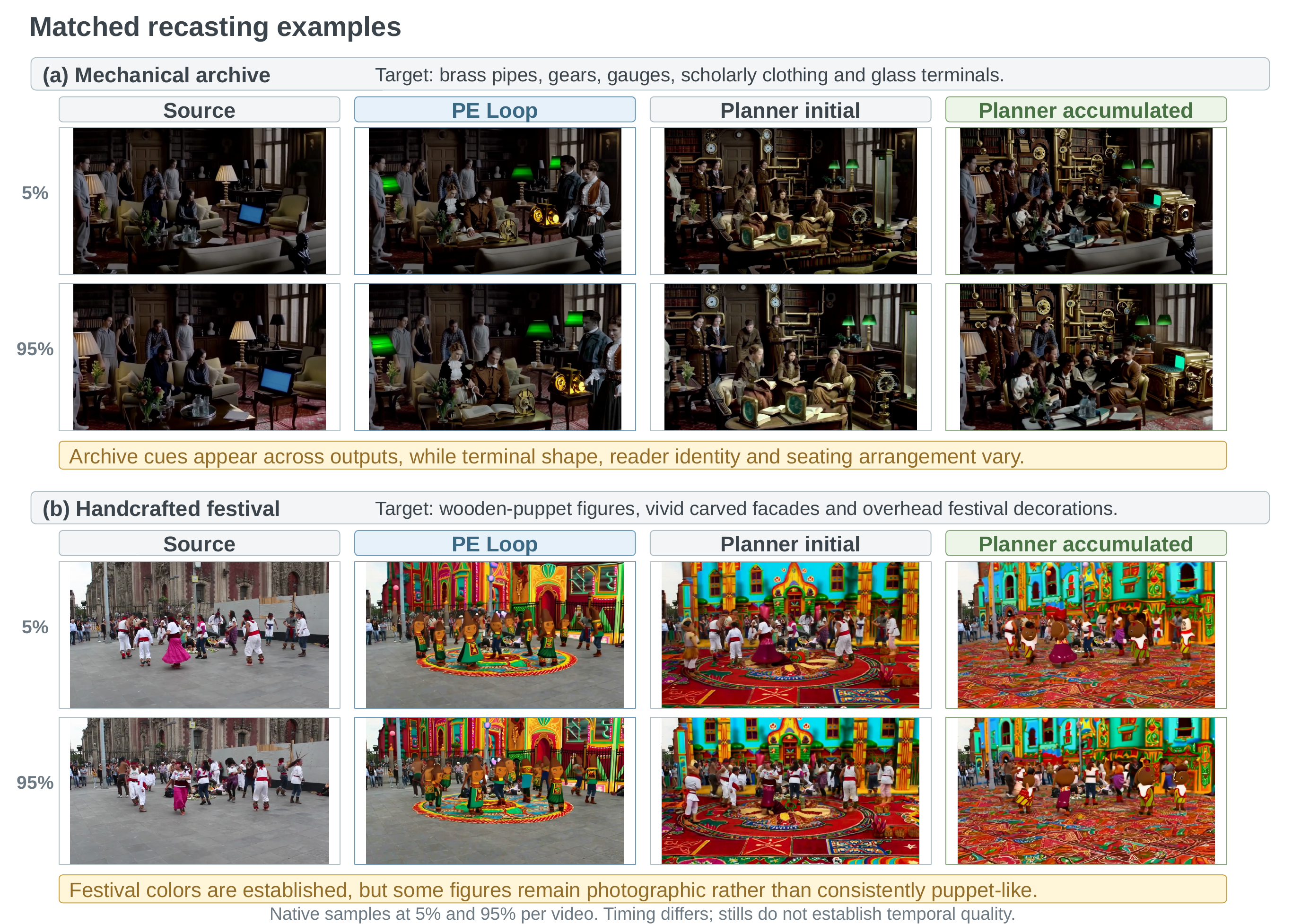}\par\smallskip
\includegraphics[width=\textwidth,trim=0 55bp 0 489bp,clip]{figures/revisedS2_matched_cases.pdf}
\caption{Mechanical archive (053) and handcrafted festival (059). Each row
compares the source, PE Loop, $E_0$, and $E_1$ at the same fraction of
each video's duration (5\% or 95\%).}
\label{fig:benchmark_case}
\end{figure*}

\noindent\textbf{Reading the examples.}
Main Figure~\ref{fig:paired-specifications} compares two requests for a
single source. Figure~\ref{fig:benchmark_case} holds the detailed request
fixed within each case and shows the selected output of each workflow. In
the library, compare the reading positions around the terminals. In the
festival, compare the dancers' material with the painted surroundings.
These details make it possible to inspect what each edit preserves and what
it transforms.

\section{Video evidence for evaluation}\label{app:video-observation}
\noindent\textbf{During editing.}
The Planner starts with six overview frames of the source and receives six
from each new result. The Critic receives six per relevant video. Both can
inspect crops and consecutive frames (2--16 per video per page, up to 32 per
call, with one frame of overlap). PE prompt revision starts from eight
source frames; its selector receives six overview frames per video and the
candidate reviews.

\noindent\textbf{After delivery.}
GLM-5.3-Flash and Claude-Sonnet-5.5 independently compare anonymized videos
with the original request and source. For a source of duration $D$ seconds,
the overview samples $\max(20,\lceil20D/6\rceil)$ evenly spaced times.
Each time maps to the nearest native frame; duplicate frames are shown once.
The source also defines consecutive intervals with boundaries every 28
frames. Each output is inspected over those same time intervals at its
native frame rate. Times beyond a shorter output are marked missing.
Crops keep their aspect ratio and are resized to a maximum long side of
960 pixels (JPEG quality 90). The judge reviews each interval separately
and uses those observations together with the overview for its final
ratings. Temporal consistency requires valid consecutive-frame evidence.

\section{Delivery and candidate selection}\label{app:statistics}
\noindent\textbf{Which version is delivered?}
A completed Planner run returns its selected candidate. If editing is
interrupted, the Planner returns the last preferred candidate, or the
latest generated one if it has yet to express a preference. PE Loop uses
its latest completed selection, falling back to the first completed candidate.
Runs with no delivered video form a separate category.

Across both editors, the Planner chose an earlier version in more than half
of normally completed runs (Table~\ref{tab:selection-counts}). This frequency
measures an editing decision. Main Table~\ref{tab:results}(a) measures its
visual outcome by comparing the selected video with the first generation.
Selections win more often than they lose in all eight editor--library--judge
comparisons. For Ditto $E_0$, GLM records 47.2\% wins, 33.3\% ties, and
19.4\% losses across 72 resolved pairs.
\begin{table}[!ht]
\centering\small
\caption{Planner deliveries per 134 conditions. Early is the subset of normal
four-candidate runs that returned $v_1$, $v_2$, or $v_3$.}
\label{tab:selection-counts}
\begin{tabular*}{\linewidth}{@{\extracolsep{\fill}}llrrrr@{}}
\toprule
Editor & Library & Normal & Interrupted & Missing & Early \\
\midrule
Bernini & $E_0$ & 129 & 2 & 3 & 66 \\
        & $E_1$ & 128 & 0 & 6 & 71 \\
Ditto   & $E_0$ & 130 & 1 & 3 & 68 \\
        & $E_1$ & 128 & 1 & 5 & 69 \\
\bottomrule
\end{tabular*}
\end{table}

\begin{table*}[t]
\centering\small
\caption{Reference anchors for five-axis video quality (0--4).
    Each axis assesses a distinct property of the delivered video;
    intermediate scores follow the rules in Appendix~\ref{app:diagnostics}.}
\label{tab:axis-anchors}
\begin{tabular*}{\textwidth}{@{\extracolsep{\fill}}p{0.12\textwidth}p{0.29\textwidth}p{0.31\textwidth}p{0.20\textwidth}@{}}
\toprule
Axis & 4: Complete & 2: Partial & 0: Absent or failed \\
\midrule
TR (Target Realization) & All central requested elements realize the target world or medium throughout the video, including subjects, props, setting, and light. & The target is recognizable; a central subject or setting stays source-like, the medium is intermittent, or changes are mainly surface treatment. & The requested target is absent or contradicted. \\
CP (Content Correspondence) & Source roles, event beats, spatial relations, and key interactions have recognizable counterparts in the target world. & The scene remains recognizable, but a central role, beat, or interaction is lost, replaced, or unreadable. & Required source correspondence is absent or contradicted. \\
TC (Temporal Consistency) & Events progress coherently; medium, identities, contacts, and motion remain consistent through the ending. & The event remains interpretable despite an important break in medium, identity, object persistence, contact, or focal clarity. & The sequence is incoherent or its central progression breaks down. \\
VI (Visual Integrity) & Subjects, objects, boundaries, and contacts are well formed for the medium, with all focal structures intact. & A consequential defect affects a focal face, hand, prop, contact, or boundary, such as melting, merging, blotches, or seams. & Essential structures collapse and the central interaction is unreadable. \\
AC (Aesthetic Coordination) & Composition, light, palette, materials, subjects, and setting form a unified design suited to the target world and moment. & Integration is weakened by conflicting visual systems, an unrequested medium mixture, generic decoration, or cluttered hierarchy. & Visual organization collapses across the central scene. \\
\bottomrule
\end{tabular*}
\end{table*}

\begin{table*}[t]
\centering\small
\caption{Bernini Direct instruction diagnostics, grouped by applicability.
Cells report means (valid conditions/67). Design and aesthetic scores use
0--4; completion, correspondence, binding, satisfaction, and damage use 0--1.}
\label{tab:direct-track-diagnostics}
\begin{tabular*}{\textwidth}{@{\extracolsep{\fill}}lcccc@{}}
\toprule
\multicolumn{5}{@{}l}{\textbf{(a) Metrics shared by both request tracks}} \\
 & \multicolumn{2}{c}{Short requests} & \multicolumn{2}{c}{Detailed requests} \\
Metric & GLM & Claude & GLM & Claude \\
\midrule
Edit completion & 0.577 (65/67) & 0.379 (66/67) & 0.711 (66/67) & 0.574 (65/67) \\
Required correspondence & 0.870 (54/67) & 0.493 (67/67) & 0.754 (63/67) & 0.576 (66/67) \\
Unrequested-damage flag & 0.719 (64/67) & 1.000 (49/67) & 0.831 (65/67) & 1.000 (38/67) \\
Major-damage flag & 0.094 (64/67) & 0.347 (49/67) & 0.123 (65/67) & 0.289 (38/67) \\
Aesthetic diagnostic mean & 2.623 (66/67) & 2.306 (67/67) & 3.049 (65/67) & 2.739 (66/67) \\
\bottomrule
\end{tabular*}

\medskip
\begin{minipage}[t]{0.49\textwidth}
\begin{tabular*}{\linewidth}[t]{@{\extracolsep{\fill}}lcc@{}}
\toprule
\multicolumn{3}{@{}l}{\textbf{(b) Short-request design}} \\
Metric & GLM & Claude \\
\midrule
Style realization & 2.203 (66/67) & 1.622 (67/67) \\
Scene-specific design & 2.288 (66/67) & 1.639 (67/67) \\
Effective creativity & 2.050 (66/67) & 1.524 (66/67) \\
\bottomrule
\end{tabular*}
\end{minipage}\hfill
\begin{minipage}[t]{0.49\textwidth}
\begin{tabular*}{\linewidth}[t]{@{\extracolsep{\fill}}lcc@{}}
\toprule
\multicolumn{3}{@{}l}{\textbf{(c) Detailed-request constraints}} \\
Metric & GLM & Claude \\
\midrule
Binding accuracy & 0.796 (62/67) & 0.542 (66/67) \\
Joint satisfaction & 0.015 (66/67) & 0.000 (66/67) \\
\bottomrule
\end{tabular*}
\end{minipage}
\end{table*}

\newpage
\noindent\textbf{Pairwise preferences.}
Each judge scores both presentation orders. Agreement yields a win, tie, or
loss; the win rates in the main paper use these resolved pairs.
For the group-weighted score, wins, ties, and losses count as 1, 0.5, and 0.
A sole delivery scores 1; when neither method delivers, both receive 0.5.
An unresolved verdict contributes the interval $[0,1]$. We average within
each related-source group, then weight the groups equally.

\section{Quality scores and instruction diagnostics}\label{app:diagnostics}\label{app:controls}
\noindent\textbf{Scoring a video.}
Judges score each axis in tenths using Table~\ref{tab:axis-anchors}. A 3
indicates minor secondary gaps; a 1 indicates superficial or fragmented
realization. A consequential central failure caps the affected axis at 2.
We average the available ratings within each source, then weight sources
with valid ratings equally.

\smallskip
\newpage
\noindent\textbf{Completing edits versus satisfying the whole request.}
On detailed requests, GLM gives Bernini Direct a mean edit completion of
0.711, but joint satisfaction is only 0.015 (Table~\ref{tab:direct-track-diagnostics}).
The latter requires every applicable edit, binding, and source correspondence
to pass without unrequested damage. Claude flags such damage in all 38
detailed outputs with a valid damage rating.

\smallskip
\begin{figure*}[!t]
\centering
\includegraphics[width=\textwidth]{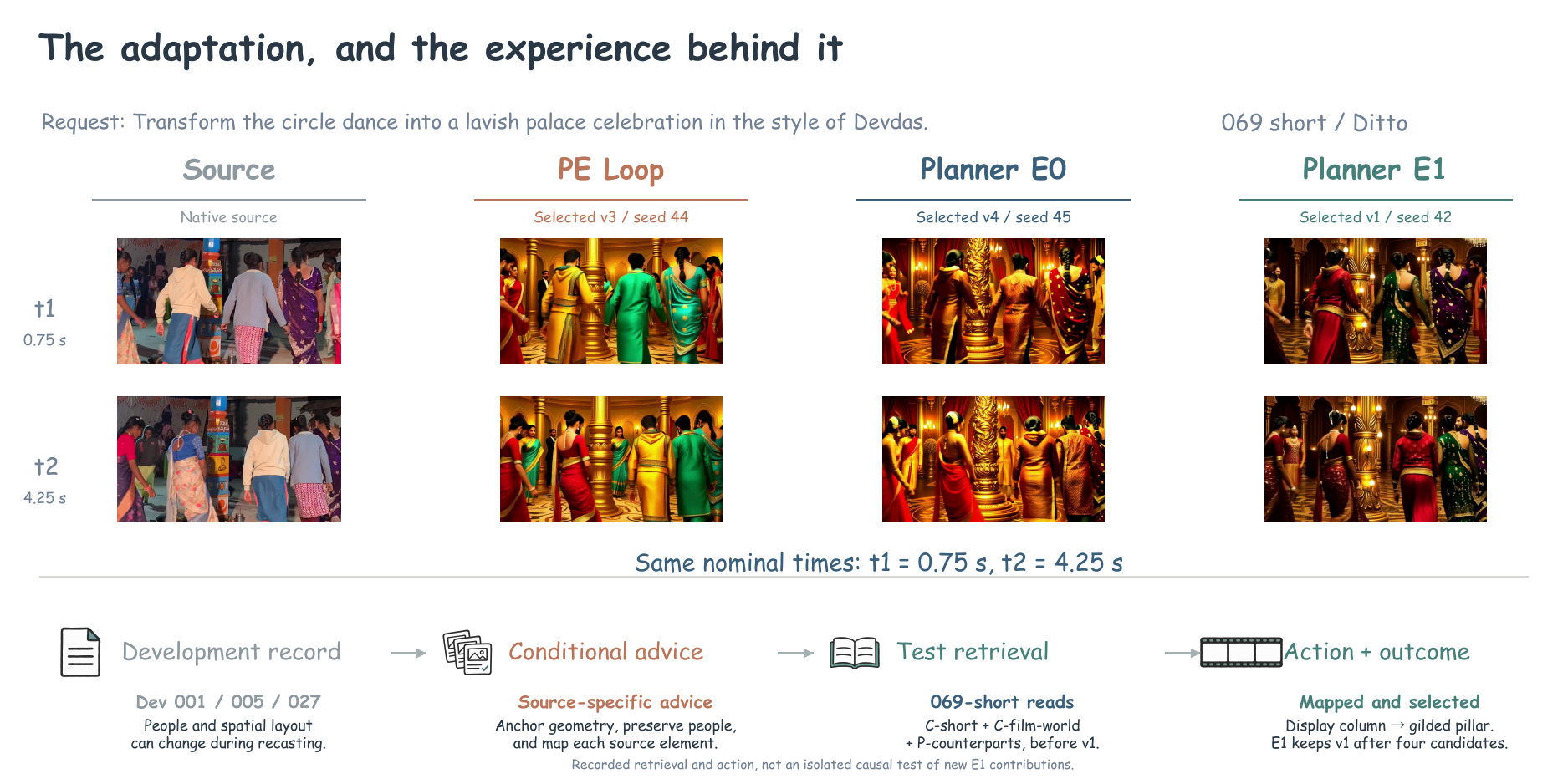}
\caption{Held-out Ditto task 069, short request. The aligned frames show the
circle dance and its palace counterparts; the trace links retrieved advice
to the first edit and the final selection of $v_1$.}
\label{fig:experience-use-trace}
\end{figure*}
\begin{table*}[t]
\centering\small
\caption{Held-out accumulated/initial comparison ($E_1/E_0$): 27 sources,
    54 conditions, and 25 groups. Final compares selected outputs and First
    compares first generations. Rates and $n$ use order-consistent pairs;
    unresolved comparisons remain in the group-weighted bounds.}
\label{tab:experience-results}

\begin{tabular*}{\textwidth}{@{\extracolsep{\fill}}lllcccc@{}}
\toprule
Judge & Editor & Stage & Win (\%) & Tie (\%) & Loss (\%) & $n$ \\
\midrule
GLM & Bernini & Final & 60.7 & 0.0 & 39.3 & 28 \\
 &  & First & 54.1 & 5.4 & 40.5 & 37 \\
 & Ditto & Final & 66.7 & 0.0 & 33.3 & 21 \\
 &  & First & 42.1 & 5.3 & 52.6 & 19 \\
Claude & Bernini & Final & 51.4 & 0.0 & 48.6 & 37 \\
 &  & First & 55.0 & 2.5 & 42.5 & 40 \\
 & Ditto & Final & 48.5 & 0.0 & 51.5 & 33 \\
 &  & First & 55.3 & 0.0 & 44.7 & 38 \\
\bottomrule
\end{tabular*}
\end{table*}

\noindent\textbf{Diagnostic denominators.}
Checks score 1 (pass), 0.5 (partial), or 0 (fail). For $m$ known checks
totaling $s$ and $q$ unknown checks, coverage is $m/(m+q)$ and the mean
lies in $[s/(m+q),(s+q)/(m+q)]$. A point score requires $m>0$ and $q=0$.
Each cell gives its valid count out of 67. The aesthetic mean requires all
five components: narrative fit, world consistency, composition and light,
color and texture, and visual hierarchy.

\section{Experience transfer on held-out tasks}\label{app:experience-case}
\noindent\textbf{What changes between $E_0$ and $E_1$?}
Both Planners start from the same model weights. The initial library $E_0$
contains 21 entries curated from development pilots. The accumulated
snapshot $E_1$ incorporates contributions from development trajectories
before either configuration is evaluated on the 54 held-out conditions.
The held-out comparison thus changes which advice the Planner can retrieve.

\newpage
Among order-consistent comparisons of selected outputs, GLM favors $E_1$ on
both editors, while Claude's preferences are close to even
(Table~\ref{tab:experience-results}). Including unresolved outcomes gives
group-weighted score bounds of $[0.35,0.77]$ and $[0.30,0.80]$ for GLM
(Bernini and Ditto), and $[0.38,0.63]$ and $[0.33,0.64]$ for Claude.
Each interval includes the equal-preference score of 0.5.

\smallskip
\newpage
\noindent\textbf{From retrieved advice to a palace scene.}
On Ditto task 069, the source dancers circle a central display. Before its
first edit, $E_1$ retrieves advice on short requests, film worlds, and
source-object counterparts. The resulting prompt maps the display to a
gilded pillar and dresses the dancers as palace guests, preserving their
positions around the center. After exploring a continuation and fresh edits
from the source, the Planner returns to this first version for delivery
(Figure~\ref{fig:experience-use-trace}).

\end{document}